\RequirePackage{snapshot}
\documentclass{article}
\usepackage{arxiv}
\usepackage{graphicx}

\title{Adaptive sampling for scanning pixel cameras}
\author{Yusuf Duman, Jean-Yves Guillemaut, Simon Hadfield}

\usepackage{subfig}
\usepackage{todo}
\usepackage{placeins}
\usepackage{amsmath}  
\usepackage{amssymb}  
\usepackage{amsfonts} 
\usepackage{mathtools}
\usepackage[nolist,nohyperlinks,]{acronym}
\usepackage{cleveref}
\usepackage[defaultlines=3,all]{nowidow}

\graphicspath{{./images/}}
\newcommand{\brackets}[3]{\ensuremath{\left#1 #2 \right#3}}
\newcommand{\mymag}[1]{\brackets{\|}{#1}{\|}} 

\newacro{sr}[$f$]{samplerate}
\newacro{a_res}[$A$]{angular resolution}
\newacro{a_vel}[$\dot{\theta}$]{angular velocity of the sensor}
\newacro{acc_angle}[$\alpha$]{Sensor acceptance angle}
\begin{document}
\maketitle
\begin{abstract}
A scanning pixel camera is a novel low-cost, low-power sensor that is not diffraction limited.
It produces data as a sequence of samples extracted from various parts of the scene during the course of a scan.
It can provide very detailed images at the expense of samplerates and slow image acquisition time.
This paper proposes a new algorithm which allows the sensor to adapt the samplerate over the course of this sequence.
This makes it possible to overcome some of these limitations by minimising the bandwidth and time required to image and transmit a scene, while maintaining image quality. 
We examine applications to image classification and semantic segmentation and are able to achieve similar results compared to a fully sampled input, while using 80\% fewer samples.
\end{abstract}
\begin{figure*}[ht]
	\centering
	\subfloat[A scanning pixel camera]{
		\includegraphics[width=0.3\linewidth]{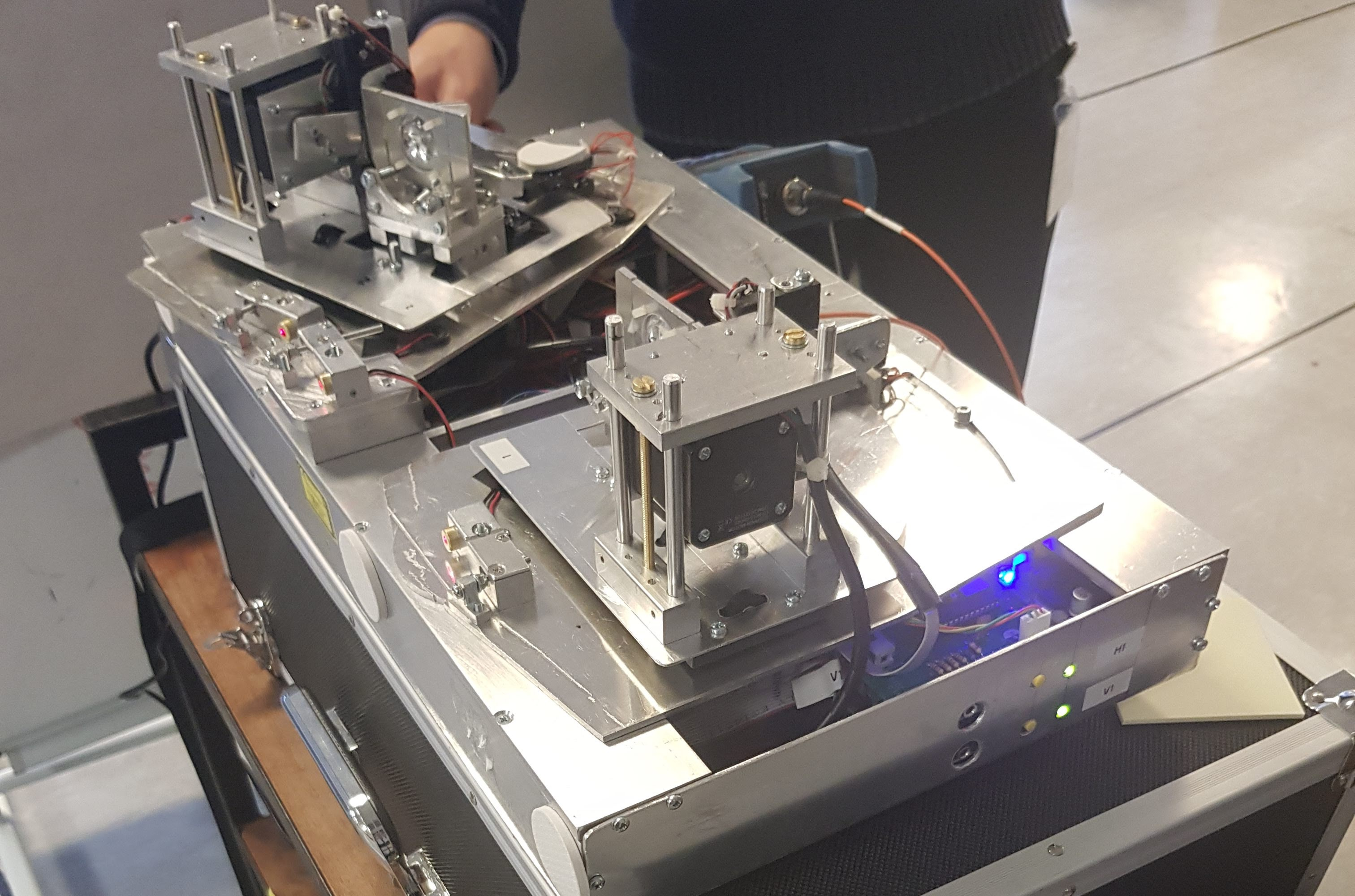}
		\label{fig:samps:spcam}
	}
	\subfloat[Full image]{
		\includegraphics[width=0.3\linewidth]{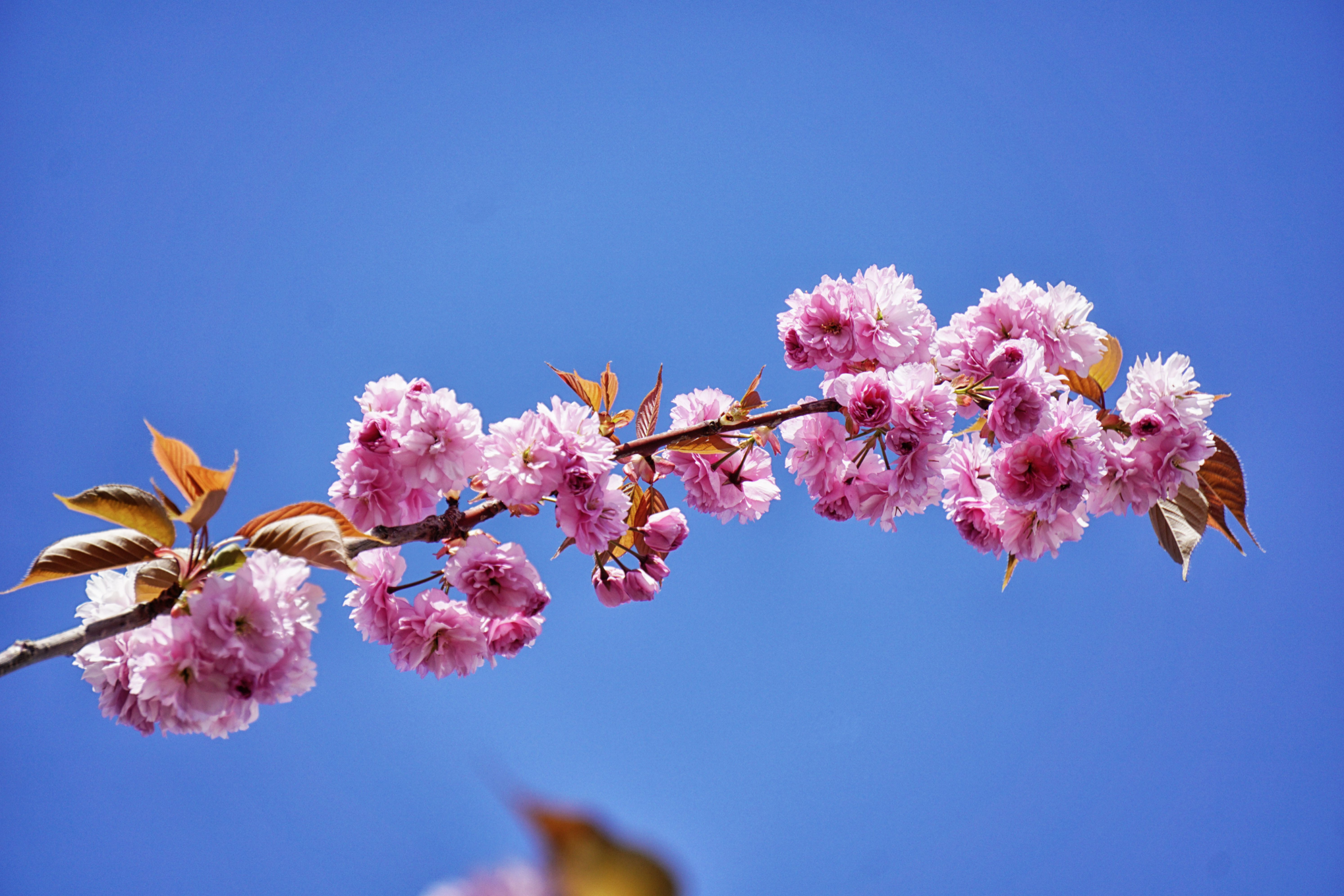}
	}
	\subfloat[SAUCE sampling]{
		\includegraphics[width=0.3\linewidth]{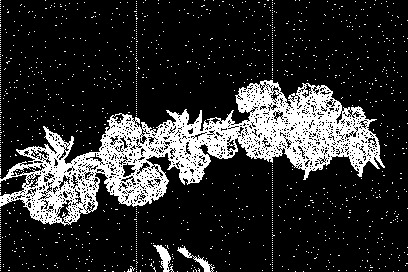}
		\label{fig:samps:samples}
	}
	\vspace{1ex}
	\caption{An example of the scanning pixel camera (\cref{fig:samps:spcam}), and the sampling map produced by the proposed technique (\cref{fig:samps:samples})}
	\label{fig:samps}

\end{figure*}
\section{Introduction}
Real world images are complex and contain a large amount of redundancy.
One of the key tasks of modern neural networks in computer vision is learning to identify the relevant details and ignore the irrelevant ones.
This has led to the development of models that explicitly learn to avoid redundant areas of the image, such as that by Yue \textit{et al.} \cite{yue2021vision}, which results in a greater accuracy on an image classification task. 
Unfortunately, although deep learning has proven very effective at this, it is invariably employed after the image acquisition.

In contrast, mammalian vision systems have a strategy to ignore redundant information at the acquisition stage.
There is a relatively small region of high density at the centre of the retina which is responsible for providing high detail, and the density of photoreceptors drops rapidly as we move away from this \cite{wandell1997foundations}.
Eye motion (saccades), focus on key details and a complex visual processing system allows a detailed image to be built from a relatively small amount of high-fidelity information.

By reducing the area of an image that we actively view, we can reduce the amount of data required to represent the image.
Taking this to the extreme, we can ask what is the least amount of image that needs to be viewed while still maintaining task accuracy.
This is useful in situations with limited bandwidth such as earth observation satellites for weather monitoring, or edge-connected internet-of-things devices.

Thanks to advances in optics and micro-actuation it has now become possible to create a camera that operates based on this principle.
This uses a single photodetector mounted at the focal point behind a moveable set of optics. 
An example of such a system can be seen in \cref{fig:samps:spcam}. 
The optics gather light that falls within a specific acceptance angle, so a sample taken at the focal point will aggregate all the information from within that acceptance angle. 
To image a scene the optics are moved across the scene and the focal point is repeatedly sampled at a frequency, \acs{sr}. 
Each sample will correspond with a different, and potentially overlapping, spatial location. 
We can treat these samples as pixels and by arranging them according to their spatial location a viewable image is formed. 

In order to scan without overlapping samples, the following condition must be met
\begin{equation}
	\acs{sr} \leq \frac{\acs{a_vel}}{\acs{acc_angle}}
	\label{eqn:max_sr}
\end{equation}
where \acs{a_vel} is the angular velocity of the sensor and \acs{acc_angle} is the acceptance angle of the optics.
Beyond this point the overlapping samples will cause a box blur of the resulting image.

Although the sensor can theoretically achieve extremely high resolutions, the limiting factor of the scanning pixel camera is the time taken to acquire, transmit and process an image, particularly if a high samplerate is being used.
Therefore there is an incentive to minimise the number of samples taken while maintaining the quality of the image. 
This will speed up image acquisition, while reducing processing and transmission times, and reduce any artefacts produced by the scene changing over time (e.g. rolling shutter artefacts).
Therefore it is desirable to reduce the samplerate for areas that require a lower resolution as it will improve the time taken to form an image without a loss of quality, similar to the function of saccades in a mammalian vision system. 
\\
In order to fully realise a saccade like system using a scanning pixel camera, it is first necessary to establish where best in the scene to take samples.
This paper explores this problem by presenting the following contributions.

\begin{itemize}
    \item A novel method called SAUCE (Sampling Adaptively Using Controlled Encoding) will be presented, which adaptively samples sequential data, particularly scanning pixel camera data.
    
    \item A novel normalisation mechanism that allows for the selection of the samplerate for a trade-off between accuracy and computational efficiency.

    \item We present an approach to allow for the integration of SAUCE in an end-to-end pipeline which includes downstream computer vision tasks, such as image classification.
    In doing so we optimise SAUCE for the computer vision task.
    This demonstrates superior performance when compared to current sampling methods when using a sampled input.
\end{itemize}

\section{Related work}
At a pixel level it is possible to see which areas of the input image caused the output.
These methods, designed to give visual explanations, make it easier for the user to understand how the network came to such a conclusion.
One of the earliest examples of this for deep learning is guided backprop  \cite{springenberg2015striving}, which looks at the products of the activations and their gradients as they flow back through the image.
More recent approaches such as GradCAM \cite{selvaraju2017grad} and AblationCAM \cite{ramaswamy2020ablation} improves on these visualisations by making them class discriminative.

These visual explanations tell us that useful information is concentrated in certain areas of the image. 
It is natural to ask whether we can discard the unimportant regions rather than simply down-weight their importance. 
Yue \textit{et al.} \cite{yue2021vision} demonstrates that this is possible using a network to iteratively shift patches toward the relevant areas of the image to provide better tokenised information to the vision transformer.
By ignoring redundant information, the network increases the accuracy of a classifier by 3.8\% and is able to reduce the number of parameter required for the task.

Compressive sensing can be used to reconstruct images with fewer samples than the Nyquist limit would require \cite{4472247}. 
Here a stationary sensor is used and samples are taken while the input is modulated using a variety of sparsely patterned masks.
This produces an under-determined linear system which can be solved to produce the image \cite{candes2006stable}. 
A more modern deep learning approach \cite{Higham2018}  uses an autoencoder to learn a fixed sampling pattern for reconstruction in a compressive sensing setup.
In contrast the work proposed here attempts to bypass the image reconstruction step entirely, and instead aims to select the best samples for each input to solve a downstream task.
Some previous work has been done directly integrating a sampled image into an action recognition pipeline \cite{kulkarni2015reconstruction}.
Here 3D MACH filters were used to condition the random patterns in a DMD array to allow for classification of the action directly from the output rather than performing a reconstruction.
Having a fixed optical system makes compressive sensing far less flexible than the approach proposed in this paper.
It is challenging or impossible to adapt camera properties like resolution and field of view on the fly.

Sparse sampling techniques are used in domains such as scanning microscopy, where imaging takes a long time to perform.
An example of  this is SLADS-Net \cite{zhang2018slads}, which takes the set of samples and their positions and gives the positions of where the next set of samples should be taken, deciding based on which samples would be most likely to reduce the distortion in the image.
This process is repeated until the desired number of samples is taken or the desired output quality has been reached.
This was able to decrease the image acquisition from hours to minutes. 
The closest work to our proposed technique are hard visual attention mechanisms such as \cite{elsayed2019saccader}, which divide the input into a series of glimpses that are independently processed before finally being combined in order to produce the final output for the whole image.
Our work differs as we estimate the usefulness of each pixel to the task network and forward only the important pixels to the task network, rather than operating on a set of features that are produced by a series of glimpses. 

\section{Methodology}
\begin{figure*}
	\centering
	\includegraphics[width=\linewidth]{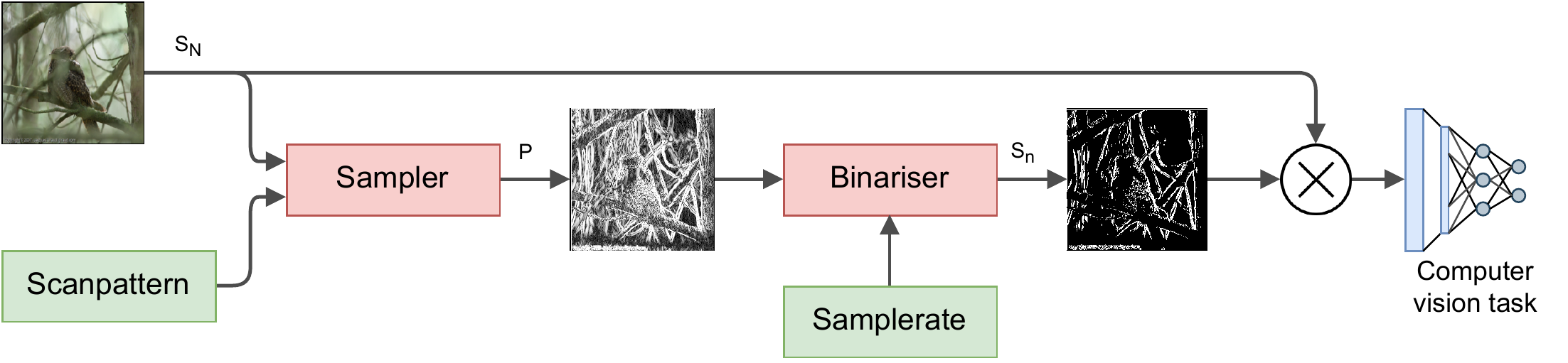}
	\caption{An overall architecture diagram of the system. 
		The sampler produces probabilities that work at all samplerates and the user selects the one desired for the downstream task.}
	\label{fig:framework}
\end{figure*}

The architecture of the proposed system is shown in \cref{fig:framework}.
The sensor images the scene following a pre-determined scan pattern.
This provides a mapping  $S_N(i): \mathbb{Z} \mapsto \mathbb{R}^{d}$ from a one dimensional input domain of potential sampling positions, $i$, to a $d$ dimensional vector space of intensity values (where the value of $d$ depends on the photodetectors used).
This makes $S_N$ the set of all possible samples for a given scene, ordered according to their sampling position.
Without loss of generality we assume for the remainder of this paper that the sensor follows a simple raster scan pattern to produce $S_N$.
During training, the sampler is given a target number of samples $n$, (or equivalent downsampling rate $n/N$, where $N$ is the total number of samples).
The system then adaptively samples the input to produce a set of samples $S_n \subset S_N$, in order to maximise the accuracy of a downstream task, $\mathcal{T}$, such as classification.


\subsection{Sampling}
For each sample, $S_N(i)$, we want an estimate of how useful that sample will be to the downstream task.
We call this estimate $P(i)$ and refer to $P(i)\forall i$ as the sample heatmap, $P$. 
Once produced, $P$ is thresholded based on the proportion of desired samples $\frac{n}{N}$, to produce a mask $I_m$.
This mask is used to select samples from $S_N$ ensuring that SAUCE only takes $n$ samples to create $S_n$, which is used as the input to the task network.

\subsubsection{SAUCE}
The SAUCE  sampler needs to operate online and be computationally efficient, we propose to determine the probability of taking each sample as, 
\begin{equation}
P(i) = 1 - e^{\frac{-D(i)}{\sigma_D^2}}
\label{eqn:prob}
\end{equation}
where $\sigma_D^2$ is the variance of the distance function, $D$, which is defined as, 
\begin{equation}
	D(i) = \alpha\Dot{\theta}(i) + \beta\Delta I(i) + \gamma
	\label{eqn:hmap}
\end{equation}
where $\dot{\theta}$ is the change in scan position, $\Delta I$ is the change in intensity since the previous sample and $\alpha,\beta$ and $\gamma$ are learnable parameters, trained using the method discussed in \cref{training_scheme}.

$\dot{\theta}$ is inspired by the angular velocity in \cref{eqn:max_sr}, as a faster change in scan position implies a faster motion, meaning the sampling rate must increase correspondingly to observe a given spatial frequency. 
It is important to note that $\theta$ corresponds to the sensor's angular position which is determined by the scan-pattern used.
Similarly, rapidly changing intensity also implies that the sampling rate must be increased to observe a given spatial frequency. 
$\Delta I$ is used to account for this.

\subsubsection{DeepSAUCE}
\begin{figure}
	\centering
	\includegraphics[width=\linewidth]{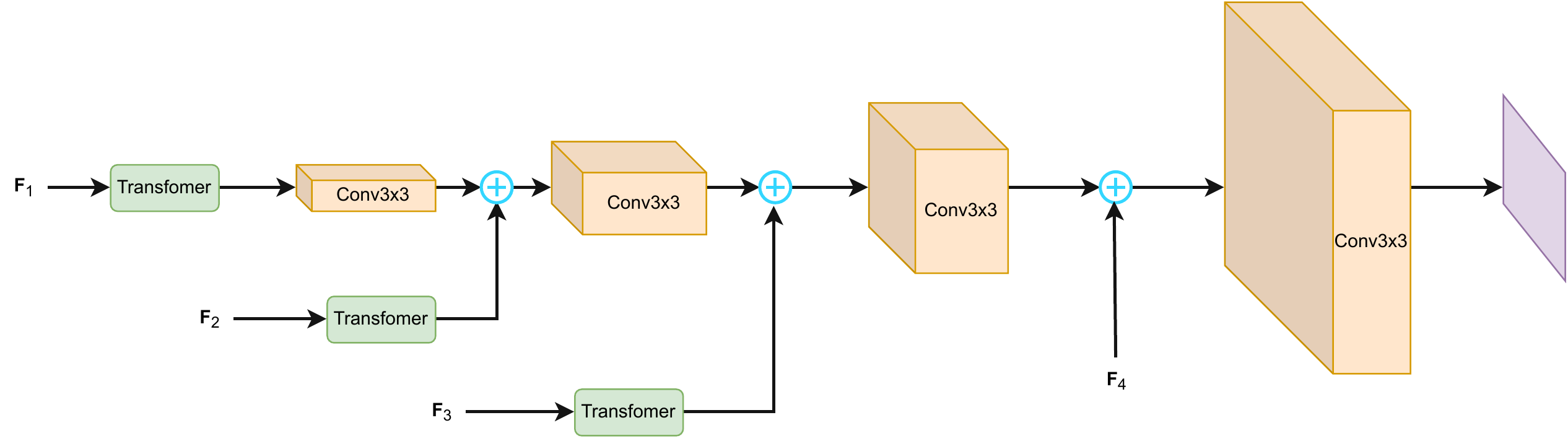}
	\caption{DeepSAUCE transformer based decoder architecture}
	\label{fig:deepsauce}
\end{figure}

The second sampling scheme we propose is a deep learning model.
We utilise an unet-like \cite{unet} architecture with transformer modules in the on the skip connections to the decoder, as shown in \cref{fig:deepsauce}.
As an encoder we use EfficientNet-b0 \cite{Mingxing19}.
A transformer is applied on the spatial positions for each of the 3 deepest feature layers.
These are then added to any prior features before having a convolution applied to them.
The last feature layer does not have a transformer applied to it and is instead directly added. 
To help maintain discrimination between the different parts of $P$, it should be ensured that the activation functions asymptotically approach a value when the input is less than $0$.
In the case of EfficentNet, the SiLU activation already achieves this, however encoder networks using activations such as ReLU will need to have them replaced.
%
When combined with the normalisation, this ensures that we do not end up with an output $P$ containing large portions clipped to $0$, which are difficult to effectively resample at different drop rates.

\subsubsection{Sample Heatmap Normalisation}
\begin{figure*}
	\centering
	\subfloat[Without normalisation]{
		\begin{minipage}{0.49\linewidth}
			\includegraphics[width=\linewidth]{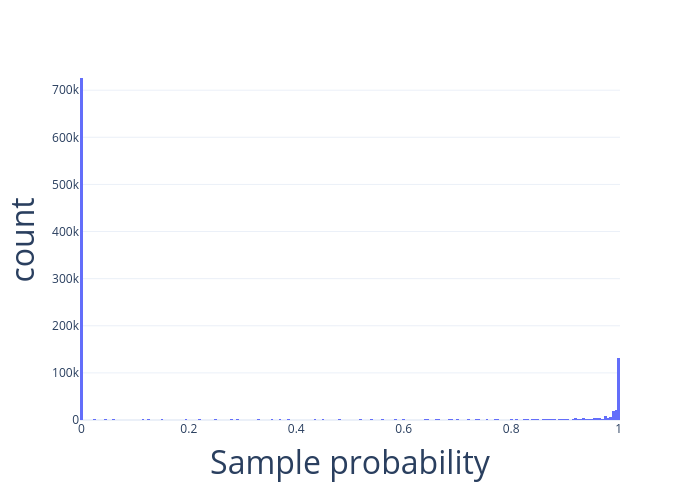} \\
			\includegraphics[width=\linewidth]{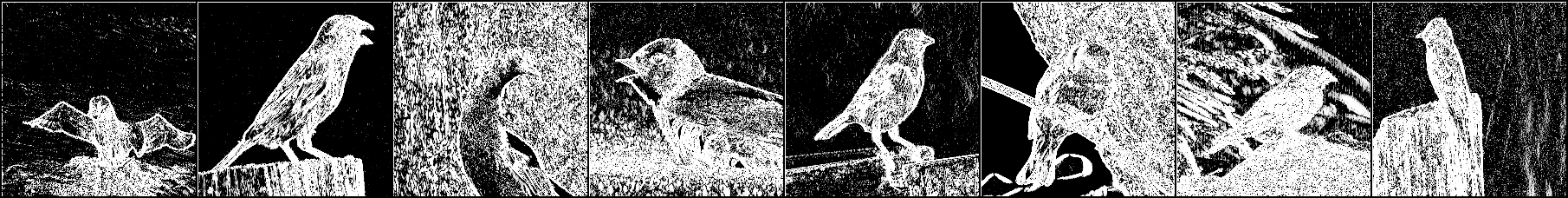}
		\end{minipage}
	\label{fig:normal:no}		
	} 
	\subfloat[With normalisation]{
		\begin{minipage}{0.49\linewidth}
			\includegraphics[width=\linewidth]{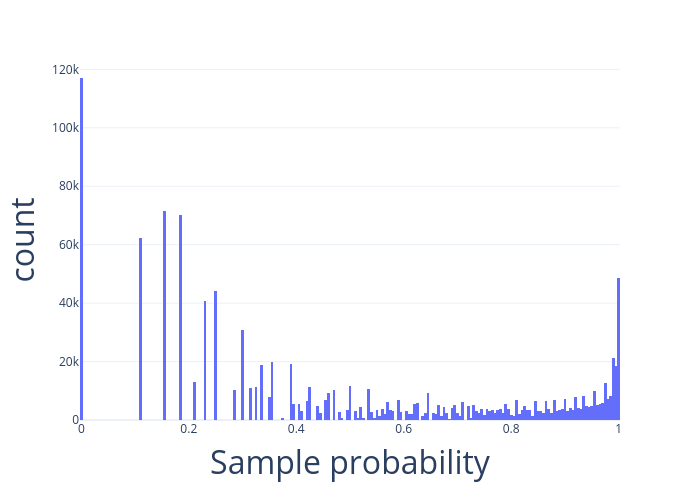} \\
			\includegraphics[width=\linewidth]{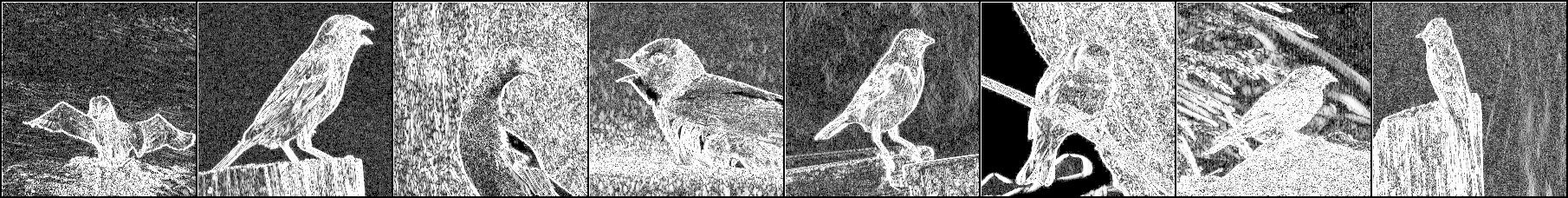}		
		\end{minipage}	
		\label{fig:normal:yes}		
	}
	\vspace{1ex}
	\caption{Distribution of sampling probabilities with and without normalisation.
		The normalisation produces a wider range of probabilities which helps with generalisation to other target samplerates at runtime.
	}
	\label{fig:normal}
\end{figure*}

If we apply no constraints to the heatmap it becomes heavily binarised (\cref{fig:normal:no}), so the sampler only operates at the one that it was trained for.
To achieve a target samplerate, $\frac{n}{N}$, without having to retrain a new network, we add the logit of the target samplerate as follows,
\begin{equation}
P = \hat{P} + \ln \left( \frac{n}{N-n} \right)
\label{eqn:scaling} 
\end{equation}
This modifies the amount of samples taken by ensuring that each sample will have a base probability corresponding to the target sample rate. 
$P$ will then increase these probabilities in important regions, and decrease in unimportant ones.  
After the application of \cref{eqn:scaling} $P$ is rescaled so the minimum value is $0$ and the maximum value is $1$.
This gives a wider range of probabilities (\cref{fig:normal:yes}) so the sampler can generalise to a wider range of sample rates, despite having been trained for one.
 
\subsection{Training scheme} \label{training_scheme}
We simultaneously train the sampler along with the downstream task in an end-to-end fashion.
The training of the downstream task is not modified other than masking the input according to $I_m$.
As we are attempting to get the network to operate on sampled images, we allow all the layers of the task network to train, so that it is able to produce features from a sampled input.
As the framework is end-to-end differentiable, the loss being applied to the task network ($L_\mathcal{T}$) will also train the sampler, allowing it to find samples specifically useful for solving the task. 
So that the desired samplerate is achieved we apply an L1 loss to the samplemap,
\begin{equation}
	L_{\mathrm{droprate}} =  \mymag{\frac{n}{N} - \frac{1}{N}\sum I_m}
	\label{eqn:drop_loss}
\end{equation}
In order to make the system vary to a wide variety of samplerates we randomly select the target samplerate for each training example.

\section{Results}
\subsection{Evaluation protocol}

\paragraph{Training}
As we train on images acquired from a normal camera, we make the assumption that we are observing static scenes and that no two samples are taken on overlapping areas.
We treat the images as if they have initially been raster scanned.
We train using the ADAM optimiser \cite{kingma2015adam} with a learning rate of 0.001. 
All models were trained for 100 epochs. 
An exponential decay of 0.98 was applied to the learning rate.

\paragraph{Metrics}
Each task is evaluated using its associated metric.
We evaluate across a range of samplerates, and plot the resulting metric against the samplerate. 
Ideally we want our system to achieve high task performance across all sampling rates when compared with the task network with an unsampled input, especially at the more challenging low sampling rates.

\subsection{Classification} \label{losses}

\newcommand{\imgbase}{graphs/cub/full/80712}
\begin{figure}[h]
	\centering
	\subfloat[Input image]{
	\includegraphics[width=\linewidth]{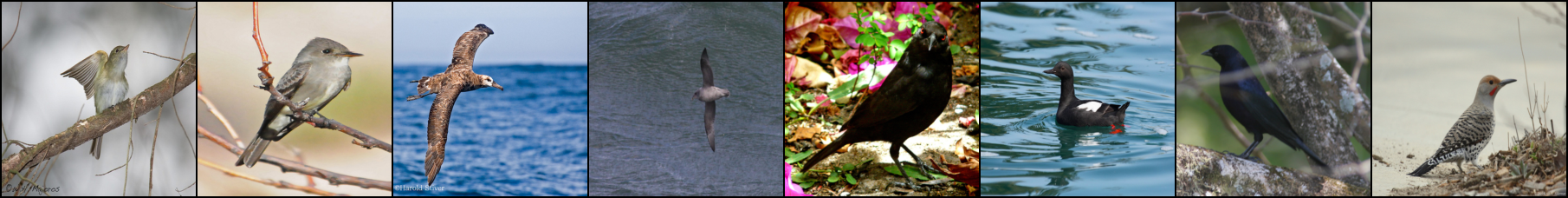}
	}

	\subfloat[Level crossing]{
	\includegraphics[width=\linewidth]{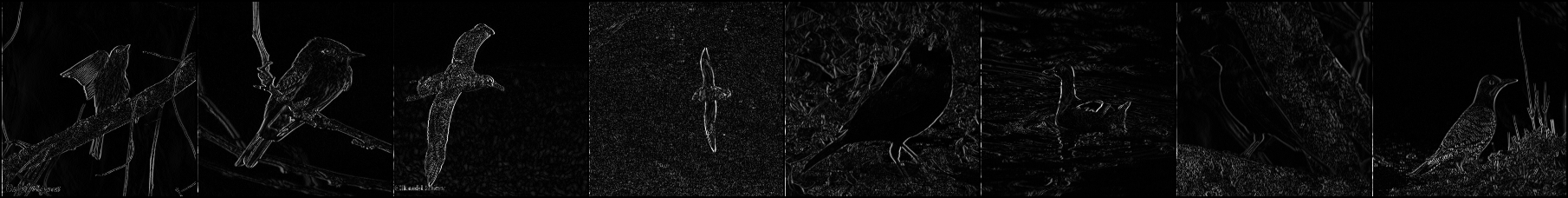}
	}
	
	\subfloat[Mixed adaptive random]{
	\includegraphics[width=\linewidth]{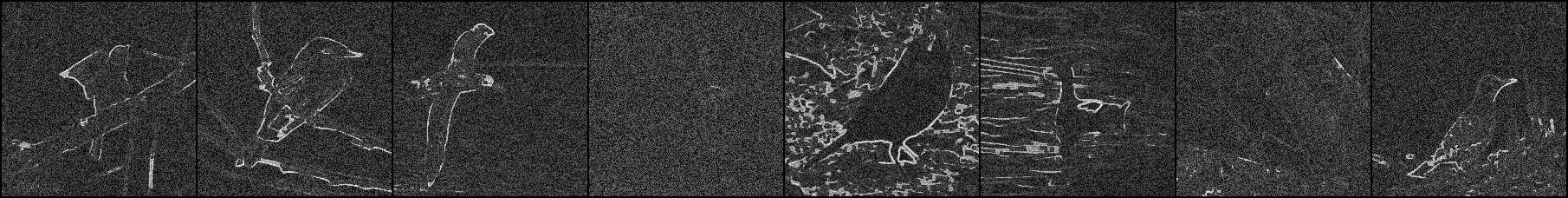}
	}

	\subfloat[SAUCE]{
	\includegraphics[width=\linewidth]{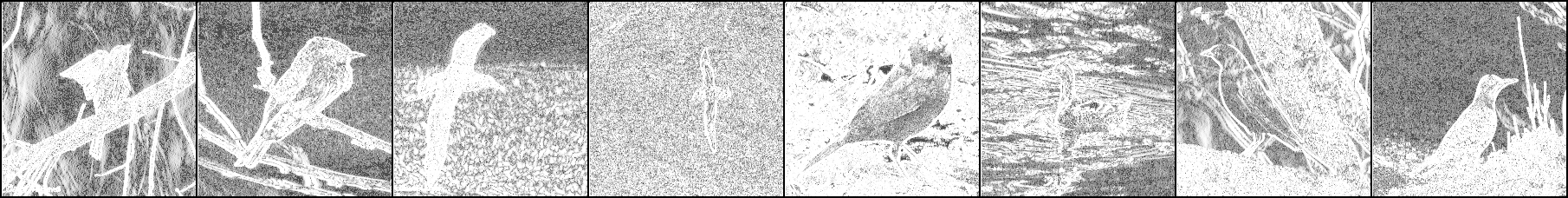}
	}

	\subfloat[DeepSAUCE]{
	\includegraphics[width=\linewidth]{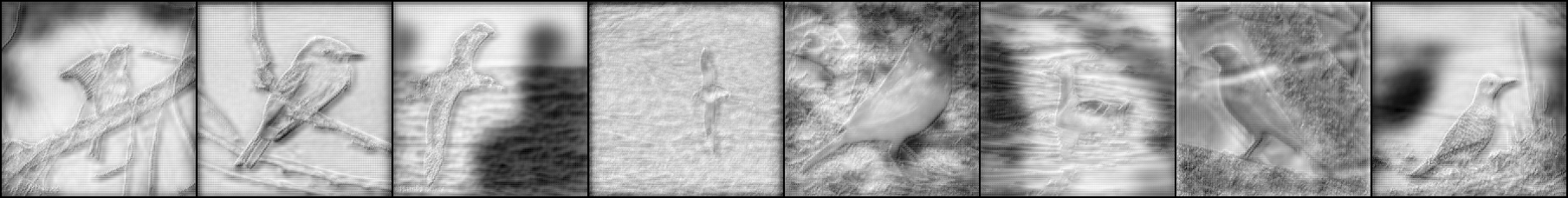}
	}
	\vspace{1ex}
	\caption{Example probability heatmaps used for classification. }
	\label{fig:examples}		
\end{figure}

\begin{figure}
\centering
\subfloat[Results of our samplers on classification of ImageNet dataset \cite{ILSVRC15}]{
	\includegraphics[width=0.47\linewidth]{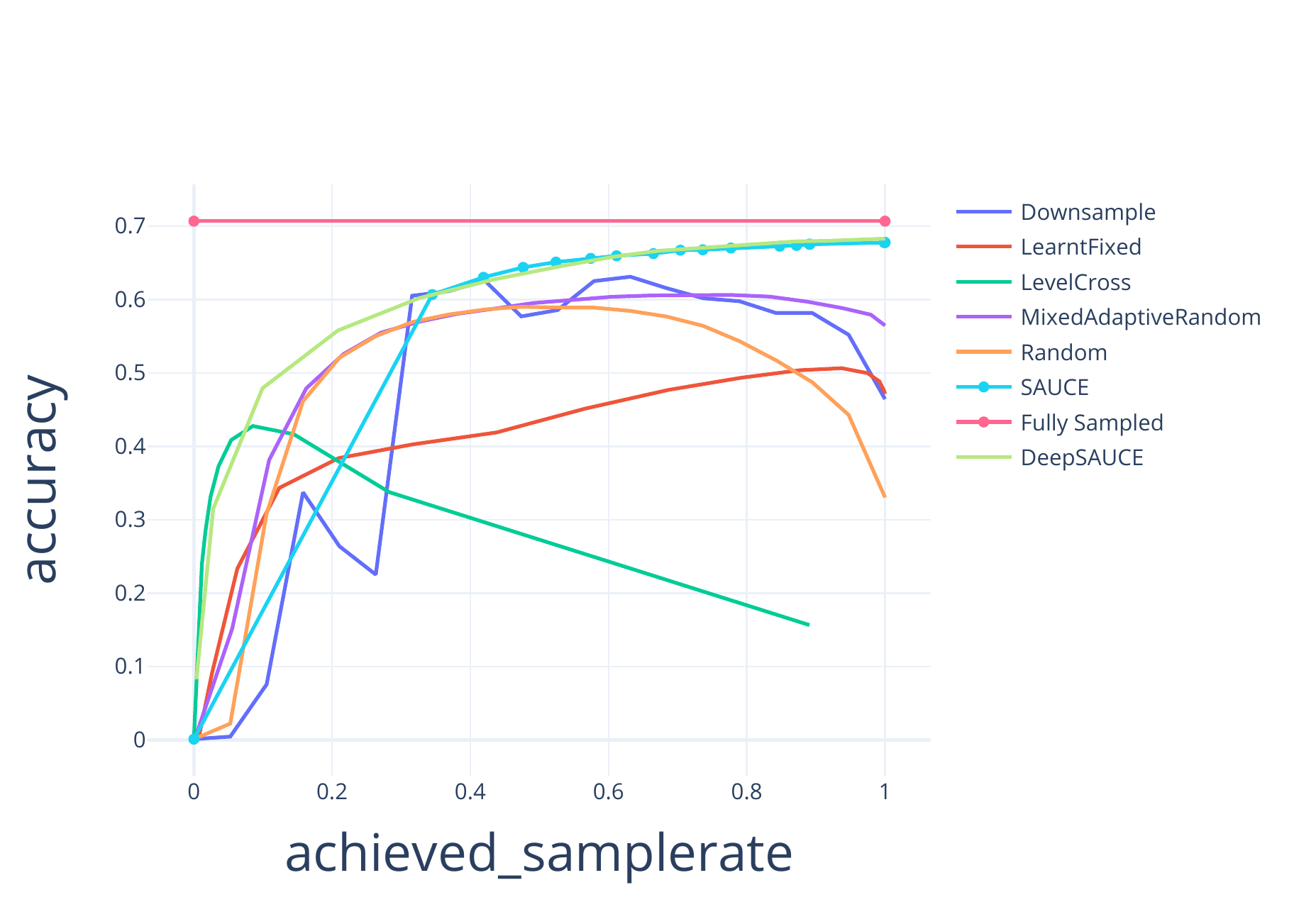}
	\label{fig:sauce_class}		
}
\hfill
\subfloat[Results of our samplers on classification of CUB200 dataset \cite{WahCUB_200_2011}]{
	\centering
	\includegraphics[width=0.47\linewidth]{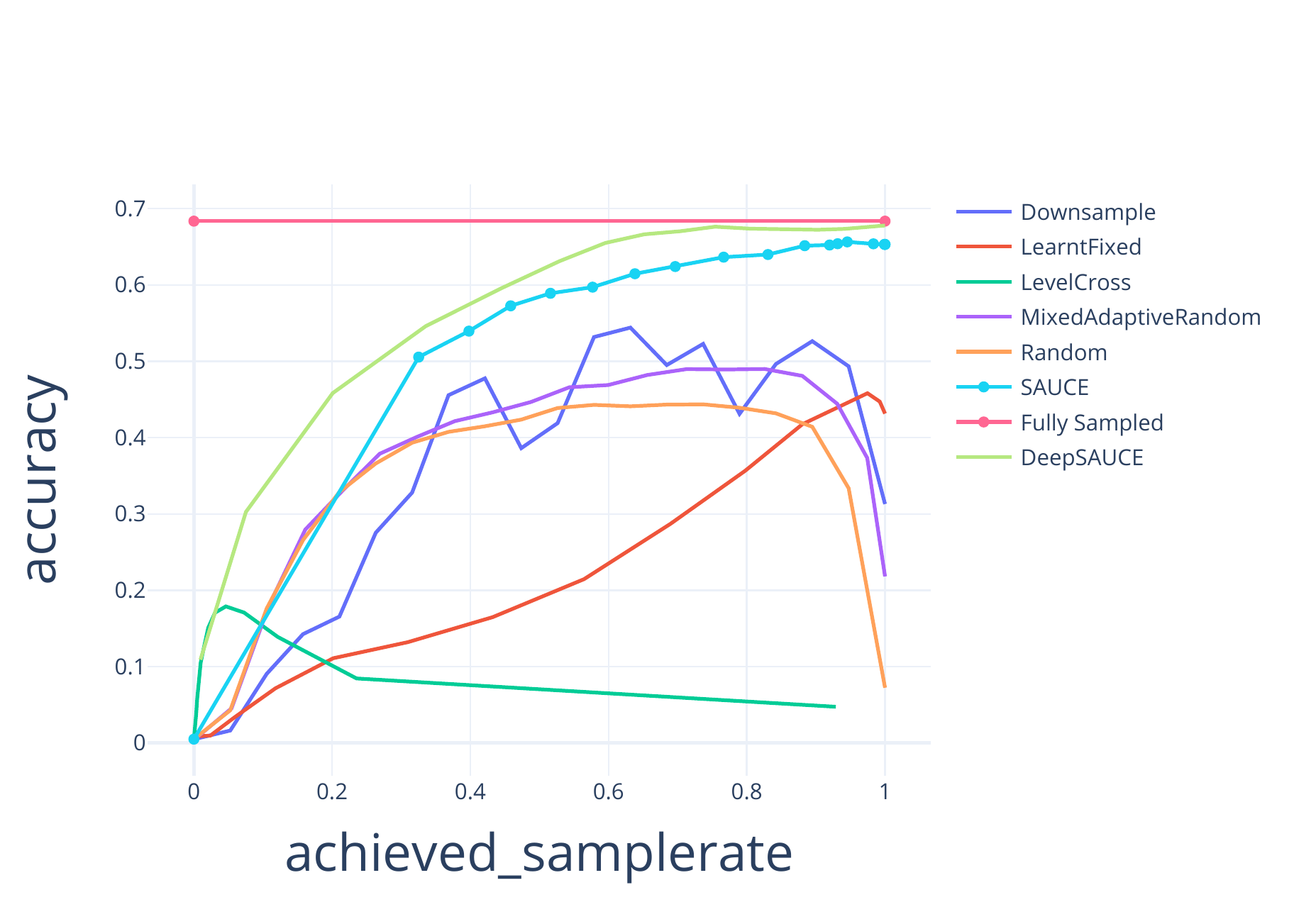}
	\label{fig:inat_class}
}
\vspace{1ex}
\caption{Classification on different datasets}
\label{fig:class}
\end{figure}

We train our samplers for the task of image classification on multiple datasets.
A pretrained EfficientNet-b0 \cite{Mingxing19} was trained with our sampling system to be used as a classifier.
From \cref{fig:class} we note that the SAUCE is able to maintain the same level of accuracy at 30\% sampling and only begin to significantly drop off after 20\%.
This shows we can ignore a significant amount of samples while still completing the classification task, which is useful for a scanning pixel sensor.
The wide performance range show that our normalisation method generalises well to a wide range of samplerates.

\paragraph{Comparison to other samplers}
In \cref{fig:sauce_class} and \cref{fig:inat_class} we compare our results to those of other samplers, as well an input where we everything is sampled.
These are uniform Downsampling, Random sampling, level-crossing sampling \cite{levelcrossed} and mixed adaptive-random sampling (MAR) \cite{yang2016high}.
From these results we see that the random, and level crossed samplers are able to achieve higher accuracies at lower sample rates on the training set but they do not generalise well as SAUCE on the test set.
SAUCE is able to perform significantly better than more recent specialised algorithms such as level-crossing and MAR as these are designed on the principle of reconstructing a full image (including redundant information) before passing it to the task network.
DeepSAUCE is further able to improve on these results and performs better at samplerates below $0.25$. This is likely due to the more discriminative heatmaps, which are shown in \cref{fig:examples}.

\paragraph{Impact of Normalisation}
\Cref{fig:norm_result} shows the effect of our normalisation.
It is clear that when it is applied the target samplerate more closely matches the achieved droprate.
This means that the normalisation is key in allowing the system to generalise to a range of samplerates even though it has been exposed to these through the samplerate scheduling during training.

\subsection{Segmentation}
\begin{figure}[h]
	\begin{minipage}{0.49\linewidth}
		\centering
		\includegraphics[width=\linewidth]{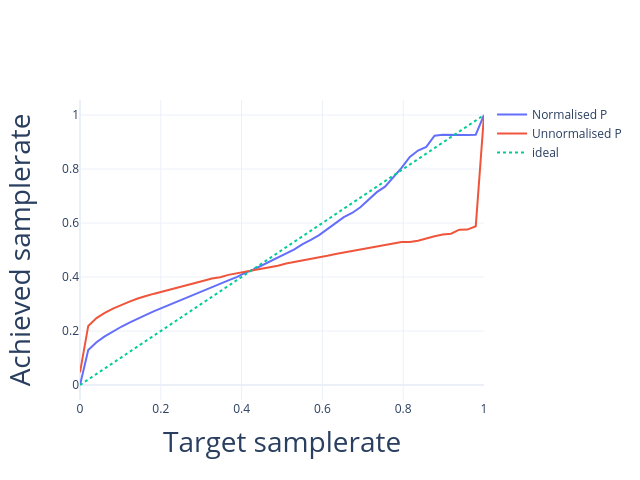}
		\caption{The effect of normalisation on SAUCE}
		\label{fig:norm_result}		
	\end{minipage}
	\begin{minipage}{0.49\linewidth}
		\centering
		\includegraphics[width=\linewidth]{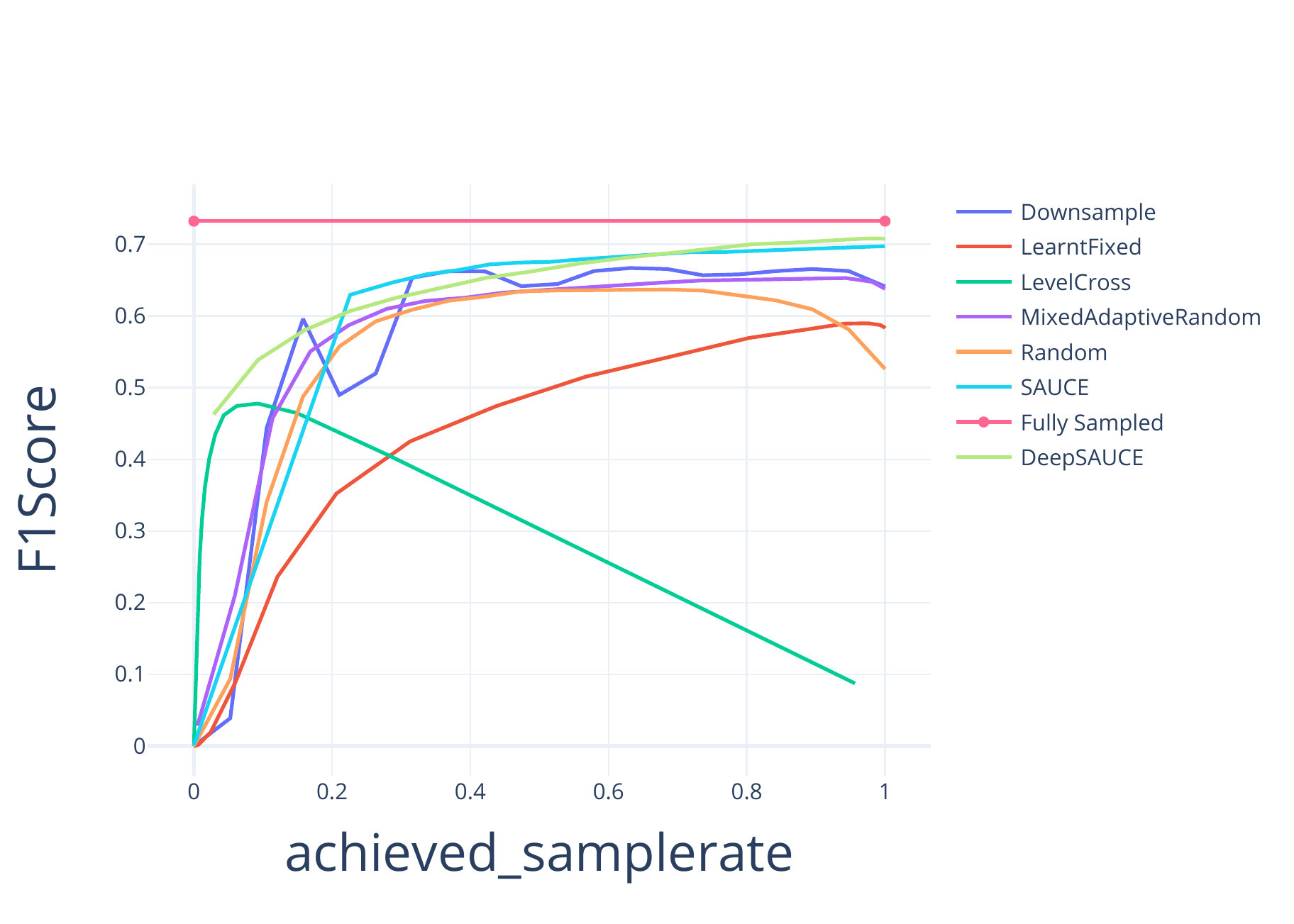}
		\caption{Results for semantic segmentation of the COCO dataset}
		\label{fig:seg_res}
	\end{minipage}
\end{figure}

As a second application we perform semantic segmentation over the COCOstuff dataset \cite{Gaidon:Virtual:CVPR2016}.
The task network used here had an encoder-decoder setup using fully convolutional resnet architecture \cite{long2015fully}.
The results for this are shown in \cref{fig:seg_res}.
These results again demonstrate that the SAUCE sampling method is better at generalising even when the result is required to form a full sized image output.
This shows SAUCE is able to operate on different tasks, however, a slightly higher samplerate is required to maintain performance, after which it significantly drops compared to the classification task.
This is understandable as establishing where the borders are is important for semantic segmentation and SAUCE puts more focus on changing areas, which are likely to form borders. 
DeepSAUCE is able to perform at lower samplerates however is also not able to do extremely low samplerates.

\subsection{Expanding to real world scenarios}
\begin{figure}
	\begin{minipage}{0.49\linewidth}
		\centering
		\includegraphics[width=\linewidth]{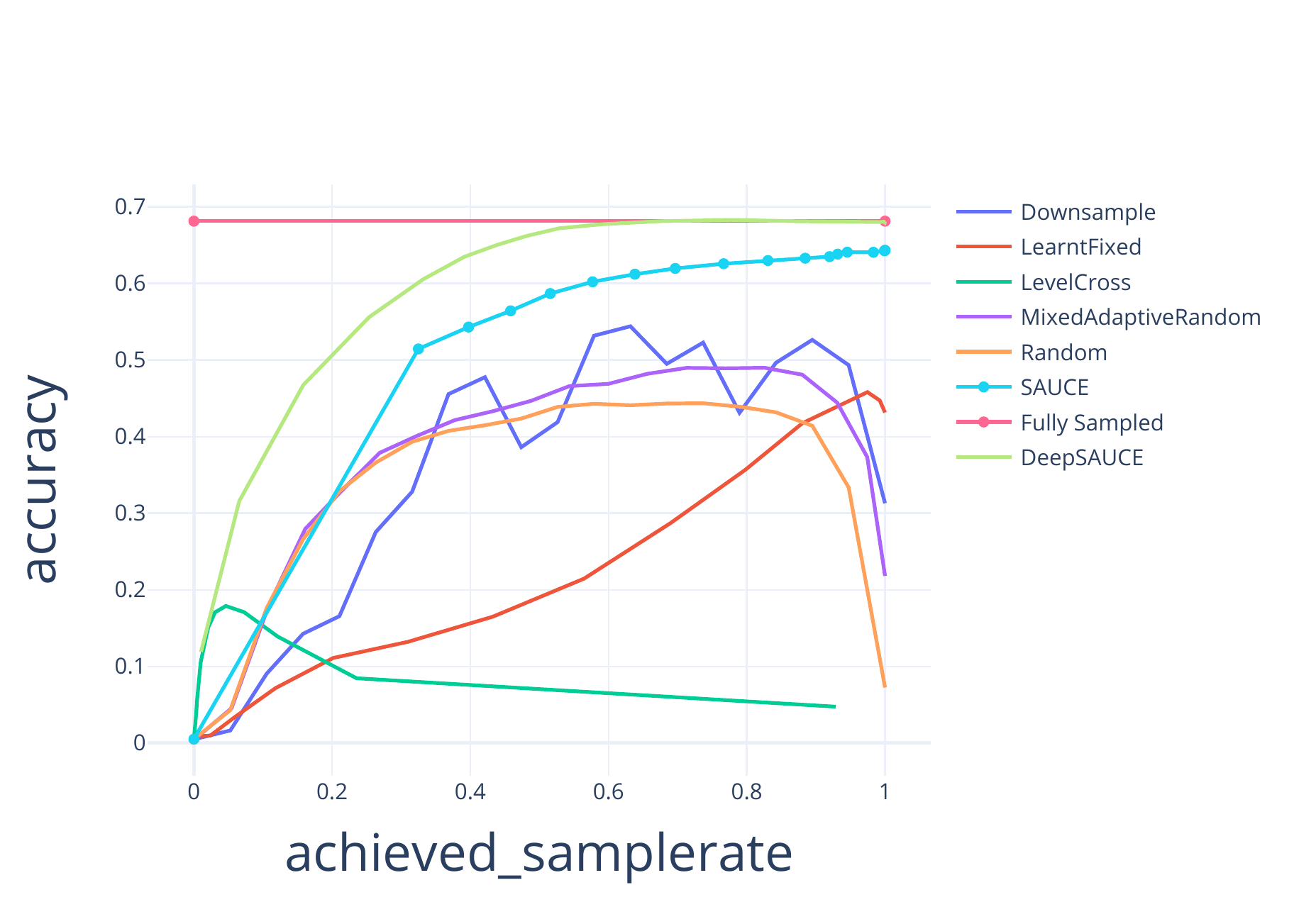}
		\caption{25\% first pass sampling}
		\label{fig:ds_25_acc}
	\end{minipage}
	\begin{minipage}{0.49\linewidth}
		\centering
		\includegraphics[width=\linewidth]{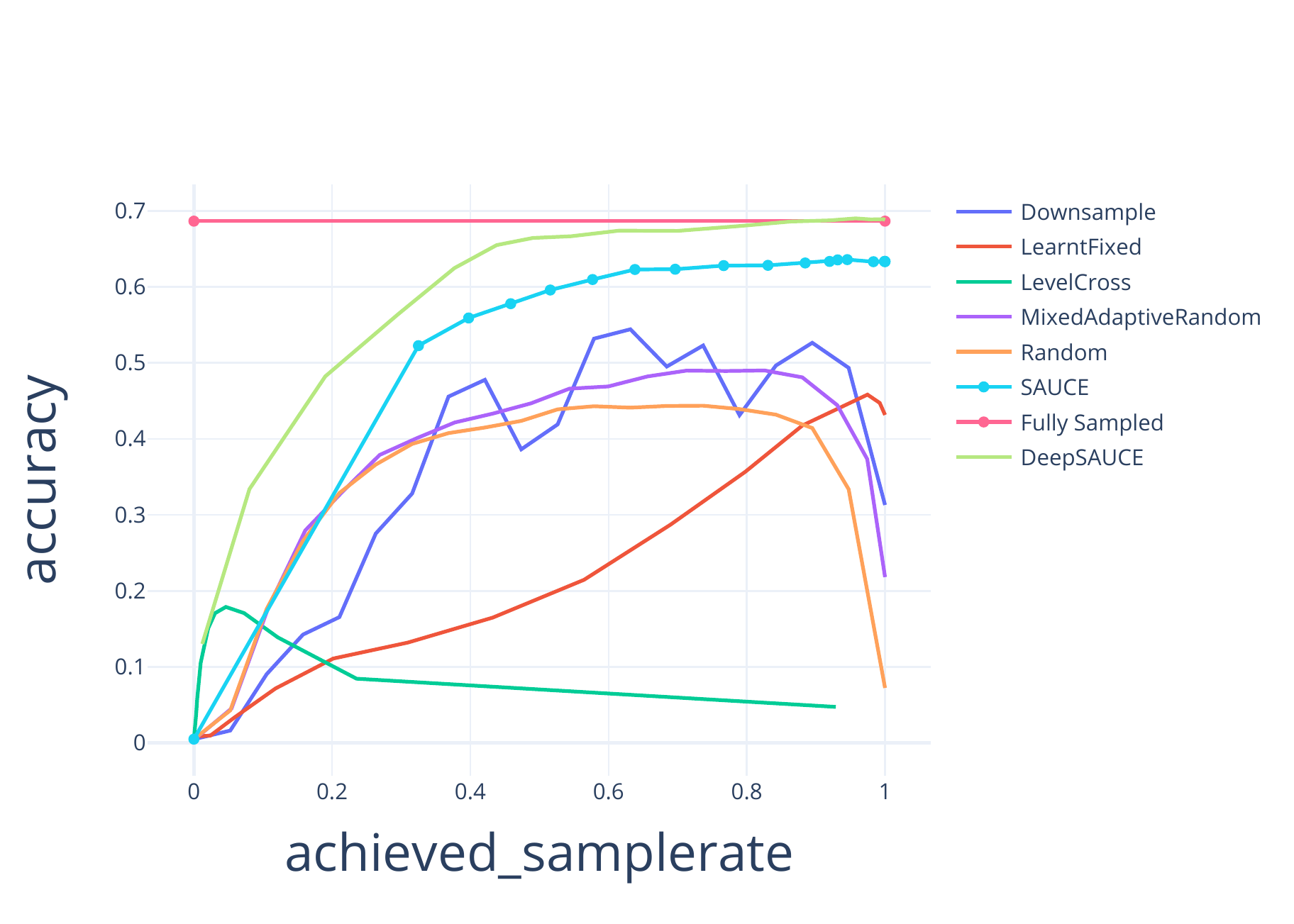}
		\caption{10\% first pass sampling}
		\label{fig:ds_10_acc}
	\end{minipage}
	
\end{figure}

A key limitation of deploying this system in the real world is that the values of all possible samples must be known, which requires that the scene be entirely imaged first.
Under this regime, the system still helps mitigate transmission bandwidth constraints, but it does not help to accelerate the imaging process.
In order to mitigate this we introduce a two stage imaging process, where an initial full scan is performed at a much lower resolution.
The resulting low-res image is used as input to produce a high-resolution samplemap to provide detail on the relevant parts.
This functionally reduces the number of samples taken despite the fact that two scans are needed.
To demonstrate this we perform classification over the CUB-200 dataset and display the results in \cref{fig:ds_25_acc,fig:ds_10_acc}. 
This shows that there is minimal loss of accuracy, even in this scenario where both the imaging time and transmission bandwidth are reduced.

\section{Conclusion}
In this paper we have demonstrated that by training a sampling scheme specific to a downstream computer vision task, it is possible to significantly reduce the number of pixels viewed by that task while maintaining accuracy.
This significantly reduces the amount of data required to meaningfully represent the input.
This is of particular interest for a scanning pixel sensor as it allows us to know where we should scan in order to build a useful image efficiently.
Future work can build on this by learning to select these samples in a predictive manner so that unused samples are never taken in the first place.


\bibliographystyle{plain}
\bibliography{sp_adaptive.bib}

\end{document}